\documentclass[10pt,twocolumn,letterpaper]{article}

\usepackage[pagenumbers]{cvpr} 

\usepackage{cite}
\usepackage{amsmath,amssymb,amsfonts}
\usepackage{graphicx}
\usepackage{textcomp}
\def\BibTeX{{\rm B\kern-.05em{\sc i\kern-.025em b}\kern-.08em
    T\kern-.1667em\lower.7ex\hbox{E}\kern-.125emX}}

\usepackage[caption=false]{subfig}
\usepackage{animate}
\usepackage{nicefrac}       
\usepackage{booktabs}
\usepackage{times}
\usepackage{microtype}    
\usepackage{array}
\usepackage[export]{adjustbox}
\usepackage{tabularx}
\usepackage{multirow}
\usepackage{multicol}
\usepackage{arydshln}
\usepackage{makecell}

\usepackage{csquotes}
\usepackage{algorithm}
\usepackage{algpseudocode}
\usepackage{url}

\usepackage[capitalize]{cleveref}
\crefname{section}{Sec.}{Secs.}
\Crefname{section}{Section}{Sections}
\Crefname{table}{Table}{Tables}
\crefname{table}{Tab.}{Tabs.}

\newcommand\Mark[1]{\textsuperscript#1}

\begin{document}

\title{Continuous Facial Motion Deblurring}
\author{
        {Tae Bok Lee}\Mark{1}, {Sujy Han}\Mark{1}, and {Yong Seok Heo} \Mark{1}\Mark{,}\Mark{2} \\
        \Mark{1}Department of Artificial Intelligence, Ajou University, South Korea\\
        \Mark{2}Department of Electrical and Computer Engineering, Ajou University, South Korea\\
        {\tt\small \{dolphin0104, tn0502wl, ysheo\}@ajou.ac.kr }
        }
\maketitle

\begin{abstract}
We introduce a novel framework for continuous facial motion deblurring that restores the continuous sharp moment latent in a single motion-blurred face image via a moment control factor. Although a motion-blurred image is the accumulated signal of continuous sharp moments during the exposure time, most existing single image deblurring approaches aim to restore a fixed number of frames using multiple networks and training stages. To address this problem, we propose a continuous facial motion deblurring network based on GAN (CFMD-GAN), which is a novel framework for restoring the continuous moment latent in a single motion-blurred face image with a single network and a single training stage. 
To stabilize the network training, we train the generator to restore continuous moments in the order determined by our facial motion-based reordering process (FMR) utilizing domain-specific knowledge of the face.
Moreover, we propose an auxiliary regressor that helps our generator produce more accurate images by estimating continuous sharp moments.
Furthermore, we introduce a control-adaptive (ContAda) block that performs spatially deformable convolution and channel-wise attention as a function of the control factor. Extensive experiments on the 300VW datasets demonstrate that the proposed framework generates a various number of continuous output frames by varying the moment control factor. Compared with the recent single-to-single image deblurring networks trained with the same 300VW training set, the proposed method show the superior performance in restoring the central sharp frame in terms of perceptual metrics, including LPIPS, FID and Arcface identity distance. The proposed method outperforms the existing single-to-video deblurring method for both qualitative and quantitative comparisons. In our experiments on the 300VW test set, the proposed framework reached 33.14 dB and 0.93 for recovery of 7 sharp frames in PSNR and SSIM, respectively.
\end{abstract}

\section{Introduction} \label{sec:introduction}

\begin{figure*}[t!]
    \captionsetup[subfloat]{labelformat=empty}
    \centering
    \includegraphics[width=0.85\linewidth]{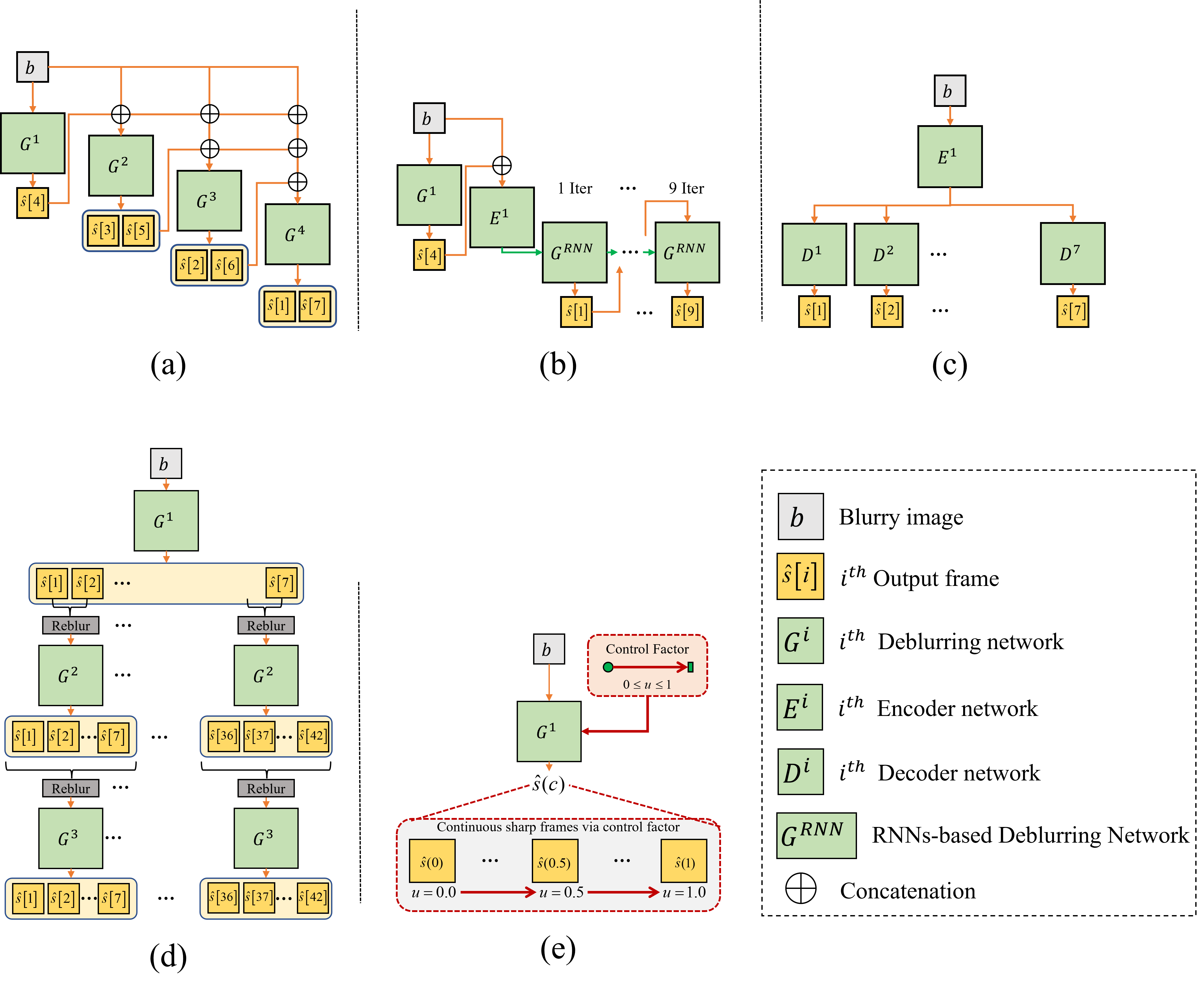}
    \caption{
    Comparison of single-to-video deblurring network architectures.
    The proposed method can restore continuous sharp motion of the face with a single network.
    (a) Jin \etal \cite{jin2018learning}, (b) Purohit \etal \cite{purohit2019bringing}, (c) Argaw \etal \cite{argaw2021restoration}, (d) Zhang \etal \cite{zhang2020every}, and (e) proposed CFMD.
    }
    \label{fig:overall_comparison_journal}
\end{figure*}

\begin{figure*}[t!]
    \subfloat[ Input blur image ]{
        \includegraphics[width=0.15\linewidth]{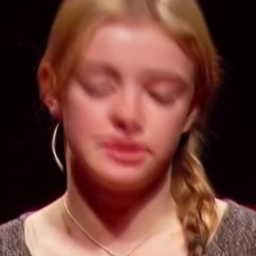}
    } \hfill \hspace*{-0.5em}  
    \subfloat[ GT (7 Fr) ]{
        \animategraphics[loop, autoplay, width=0.15\linewidth]{5}{images/comparisons_video/009_blur007_000044/gt/frame-}{0}{6}
    }\hfill \hspace*{-0.5em}  
    \subfloat[ Jin \etal \cite{jin2018learning} (7 Fr) ]{
        \animategraphics[loop, autoplay, width=0.15\linewidth]{5}{images/comparisons_video/009_blur007_000044/jin/frame-}{0}{6}
    } \hfill \hspace*{-0.5em}   
    \subfloat[ GT FMR (7 Fr)  ]{
        \animategraphics[loop, autoplay, width=0.15\linewidth]{5}{images/comparisons_video/009_blur007_000044/gt_FMR/frame-}{0}{6}
    }\hfill \hspace*{-0.5em}  
    \subfloat[ Ours (7 Fr) ]{
        \animategraphics[loop, autoplay, width=0.15\linewidth]{5}{images/comparisons_video/009_blur007_000044/cfmd_mimo_07/frame-}{0}{6}
    }\hfill \hspace*{-0.5em}    
    \subfloat[ Ours (51 Fr) ]{
        \animategraphics[loop, autoplay, width=0.15\linewidth]{5}{images/comparisons_video/009_blur007_000044/cfmd_mimo_51/frame-}{0}{50}
    }\hfill \hspace*{-0.5em}
    \caption{
    Exemplar deblurring results. \enquote{GT} denotes the ground-truth sharp frames in 300VW dataset \cite{shen2015first}. \enquote{$\#$ Fr} in parentheses denotes the number of frames. The results in (e) and (f) denote the outputs of the same network. By adjusting the control factor value, our single network can restore any number of sharp movements from a given blurry face image. \textit{This figure contains videos that are best viewed using Adobe Reader.}
    }
    \label{fig:intro}
\end{figure*}

Facial motion deblurring for a single image is a specific but critical branches of image deblurring, aimed at restoring a sharp image latent in a motion-blurred face image.
Besides being visually unpleasant, blurry face images also degrade the performance of many facial-related computer vision tasks such as face detection \cite{sun2013deep, zhang2016joint, saeed2021robust}, face recognition \cite{wang2018cosface, deng2019arcface}, facial emotion recognition \cite{yang2018facial, zhang2018facial}, and face medical image segmentation \cite{sanjar2020improved}. 
Therefore, face deblurring studies in computer vision and image processing have received much attention.

Recently, deep neural networks (DNNs) have become widespread in image restoration fields \cite{kim2016accurate, zhang2017beyond, Nah_2017_CVPR, din2020lightweight}. Among them, it has been achieved remarkable success in single image face deblurring \cite{shen2018deep, chrysos2017deep, chrysos2019motion, shen2019multi, yasarla2020deblurring, lee2020progressive, jung2022deep}.
Most of these methods recover only a single sharp image from a motion-blurred facial image.
However, motion-blurred images are the integration of continuous sharp moments during the exposure time \cite{hirsch2011fast, Nah_2017_CVPR}.
Thus, recovering such aggregated sharp moments from the blurred image can be considered the ideal goal of single image deblurring.

Several methods \cite{jin2018learning, purohit2019bringing, zhang2020every, argaw2021restoration} have been proposed to restore sharp sequences from a blurry image.
However, most of these methods have several drawbacks.
First, the temporal ordering problem is extremely challenging, because it is difficult to uniquely define the temporal order of the motion of an object in a blurry image \cite{jin2018learning, purohit2019bringing, argaw2021restoration}. 
For this reason, most existing methods fail to extract the accurate temporal order. 
This temporal ambiguity of the underlying motion in blurry images remains unsolved issue \cite{argaw2021restoration}. 
Second, as shown in \cref{fig:overall_comparison_journal}, most existing models aim to only restore fixed frames, owing to architectural design or training strategies.
Jin \etal \cite{jin2018learning} proposed a cascaded architecture consisting of four deblurring networks. As depicted in \cref{fig:overall_comparison_journal}a, each network is assigned to restore neighboring frames using the outputs from the previous networks. Thus, this method requires a large number of networks according to the number of output frames to be extracted.  
Purohit \etal \cite{purohit2019bringing} proposed the using a recurrent neural network (RNN) so that they can handle various numbers of frames without architectural changes (\cref{fig:overall_comparison_journal}b). They first extracted the middle frame using a pre-trained deblurring network and extracted nine frames using an RNN.
However, their model is fixed to restore the entire sequence with nine frames, which is the predefined number of iterations of the RNN in the training phase.
Argaw \etal \cite{argaw2021restoration} proposed a single encoder-multiple decoder architecture trained in a single training step. However, as shown in \cref{fig:overall_comparison_journal}c, this architecture requires as many decoders as output frames.
Recently, Zhang \etal \cite{zhang2020every} have shown promising results by restoring 42 frames from a blurry image. They trained three generative adversarial networks (GANs) by repeating the reblurring and deblurring processes (\cref{fig:overall_comparison_journal}d). However, they restore a fixed number of frames and require multiple training steps.

To address the problems described above, as shown in \cref{fig:overall_comparison_journal}e, we propose a facial motion-based reordering (FMR) process and a continuous facial motion deblurring network based on GAN (CFMD-GAN), a novel framework for restoring continuous moment latent in a single motion-blurred face image with a single training stage. 

To alleviate the difficulty of resolving temporal ambiguity, we estimate the reordered frames instead of estimating the frames in the original temporal order. 
To this end, we apply a facial motion-based reordering (FMR) process, which reorders frames in the dataset based on the position of the left eye in the face (\eg from top-left to right-bottom position) \cite{sun2019fab}.
This reordering process helps the network stabilize training. 

On the other hand, we introduce CFMD-GAN that restores sharp moments by varying the continuous moment control factor to estimate frames under continuous scenario.
This approach is primarily inspired by conditional GANs (cGANs) \cite{mirza2014conditional, odena2017conditional, miyato2018cgans, brock2018large, zhang2019self}, which are effective for training generators to synthesize diverse and realistic data conditioned on interpretable information, such as class labels.
In our case, a single image deblurring network serves as the generator, and the conditional information for sharp image generation is the moment control factor.
However, we have found that there are two main challenges in effectively incorporating cGANs into a single image deblurring framework.
\textbf{First}, most existing cGANs are primarily developed for image synthesis conditioned on \textit{discrete labels} (\eg class labels) \cite{ding2021ccgan}.
In contrast, we aim to restore the output images conditioned on the \textit{continuous control factor}.
Unlike most cGANs \cite{odena2017conditional, gong2019twin, kang2020contragan, kang2021rebooting} that use an auxiliary classifier for discrete class labels, we propose an auxiliary regressor to estimate the continuous control factor. It allows the proposed deblurring network to learn the image deblurring as a function of the continuous control factor. 
\textbf{Second}, an effective network module is required to apply the control factor into the deblurring network.
Most existing single image deblurring approaches directly learn image-to-image mapping functions without the use of control factor.
Recently, DNNs-based controllable image restoration models \cite{he2020interactive, kim2021searching, cai2021toward} have been extensively studied. Generally, these methods use a channel-wise attention module as a function of the control factor to resolve the Gaussian blurs and noise in static scenes.
However, spatially-variant blurs with dynamic scenes must be considered.
To this end, we present a control-adaptive (ContAda) block to effectively incorporate a control factor into recent deblurring architectures.
The proposed block learns the modulation weights using a spatially deformable convolution and channel-wise attention as functions of the control factor. 

Extensive experiments show that the proposed CFMD-GAN restores continuous sharp moments latent in a blurry face image using a single network and a single training process. \cref{fig:intro} exemplifies our results, and compares our method with previous method \cite{jin2018learning}.

The main contributions of this study are summarized as follows.
\begin{itemize}
    \item We introduce the FMR process to stabilize the network training. It allows the network to utilize rich and accurate information of the ground-truth frames corresponding to the control factor during training.
    \item We propose a CFMD-GAN for continuous facial motion deblurring that restores continuous sharp frames latent in a single motion-blurred face image via a moment control factor.
    \item We present a ContAda block to learn the feature modulation weights of the deblurring network using spatially deformable convolution and channel-wise attention as functions of the control factor.
\end{itemize}

\section{Related Works} \label{sec:related_works}
In this section, we briefly review recent single image deblurring methods and conditional GANs, which are closely related to the present work.

\subsection{Single Image Deblurring}
Traditionally, the motion-blur process is formulated as the accumulation of continuous sharp moments that occur during exposure \cite{hyun2013dynamic, Nah_2017_CVPR}.
By mimicking this, large-scale deblurring datasets \cite{Nah_2017_CVPR, noroozi2017motion, su2017deep, shen2019human} have been proposed by synthesizing a blurry image by averaging consecutive sharp frames. By leveraging such datasets, DNNs-based methods have become widespread for single image deblurring.
In the following, we introduce existing DNNs-based single image deblurring methods into three categories.

\subsubsection{Singe-to-Single, General Deblurring}
Single-to-single image deblurring aims to restore a single sharp image when a blurry image in a general scene is given.
Earlier studies \cite{chakrabarti2016neural, sun2015learning, gong2017motion} estimated the blur kernel using DNNs and obtained the resulting image using deconvolution methods.
Chakrabarti \etal \cite{chakrabarti2016neural} proposed a network that predicts the complex Fourier coefficients of a deconvolution filter and applies the predicted deconvolution filter to the input patch.
Sun \etal \cite{sun2015learning} proposed a deep learning approach that estimates motion blur kernels from local patches using a Markov random field model.
Gong \etal \cite{gong2017motion} developed a DNN to predict the motion flow from blurred images, which was used to recover deblurred images.
Without estimating the deconvolution kernel, Nah \etal \cite{Nah_2017_CVPR} utilized a coarse-to-fine network to directly restore a sharp image using their synthesized large-scale dynamic scene blur dataset.
Following the success of \cite{Nah_2017_CVPR}, variants of coarse-to-fine networks have been proposed, such as multi-recurrent networks \cite{tao2018scale, park2020multi}, multi-patch networks \cite{Zhang_2019_CVPR} and efficient multi-scale networks \cite{cho2021rethinking}.
Concretely, Tao \etal \cite{tao2018scale} designed a scale-recurrent network that shares network parameters across scales.
Zhang \etal \cite{Zhang_2019_CVPR} cascaded a multi-patch network to restore sharp images based on different patches.
In addition, Cho \etal \cite{cho2021rethinking} reduced computational costs by utilizing a U-Net \cite{ronneberger2015u}-based architecture that exploits multi-scale features extracted from an input image and outputs. 

\subsubsection{Singe-to-Single, Face Deblurring}
Face deblurring is a domain-specific task of single image deblurring that aims to obtain a sharp face from a blurry face image. 
Most existing methods have been studied in a manner that utilizes strong prior knowledge of the face, such as reference faces \cite{pan2014deblurringface, grm2019face}, face landmark \cite{chrysos2017deep, chrysos2019motion}, face sketches \cite{linlearning}, multi-task embedding \cite{shen2019multi}, 3D face models \cite{ren2019face}, facial parsing maps \cite{shen2018deep, yasarla2020deblurring, lee2020progressive} and deep feature priors \cite{jung2022deep}. 
Specifically, Shen \etal \cite{shen2018deep} proposed to estimate the facial parsing map from the blurry face and then utilize it for restoring the sharp image.
To avoid side effects caused by incorrect parsing maps, Yasarla \etal \cite{yasarla2020deblurring} utilized an uncertainty-based multi-stream architecture.
Lee \etal \cite{lee2020progressive} proposed restoring the face progressively from large components, such as skin, to small components, such as the eyes and nose.
More recently, Jung \etal \cite{jung2022deep} utilized the rich information of feature maps extracted from a pre-trained deep neural network on the face. 

However, all single-to-single deblurring methods, including the general and facial image domains, focus on restoring only one of the many moments accumulated in the blurred image. Unlike these methods, the proposed method restores various numbers of moments from a blurred image.

\subsubsection{Singe-to-Video, General Deblurring}
Instead of restoring a single output image, single-to-video deblurring is to predict multiple sharp frames from a single blurred image. In the pioneering work of Jin \etal \cite{jin2018learning}, a sequentially cascaded architecture consisting of multiple networks trained with the corresponding number of training steps was utilized. In their method, each network is assigned to predict pre-specified frames among all sharp frames. Thus, this method requires changing the number of networks based on the desired number of output frames and training them from scratch. Purohit \etal \cite{purohit2019bringing} proposed a recurrent neural networks (RNNs)-based method trained with two stages. In the first stage, they trained a video autoencoder to learn the motion and frame generation from sharp frames. It addresses the problem of the number of network scales with respect to the number of output frames. However, they still have to be trained anew each time the number of output frames changes. 
The method proposed by Zhang \etal \cite{zhang2020every} was one of the first attempts to restore continuous frames. Their method extracts a total of 42 sharp frames from a blurry image by cascading three GANs trained in three stages.
However, this approach is limited to restoring a fixed number of frames.
Instead of training the entire model in multiple stages, Argaw \etal \cite{argaw2021restoration} proposed a single framework that can be trained in an end-to-end manner.
They proposed a feature transformer network consisting of a single encoder and multiple decoders, where each decoder was specified to output a specific frame. 
Thus, this method still requires changing the number of decoders when the number of output frames changes.

In short, existing studies are inherently limited in restoring only a fixed number of frames, owing to their rigorous architectural design or training strategies. In contrast, the proposed method differs in that 1) it restores continuous sharp frames beyond a fixed number, 2) a single deblurring network with a single training step is utilized, and 3) the proposed method can be trained in an end-to-end manner.

\subsection{Conditional Generative Adversarial Networks}
Generative Adversarial Networks (GANs) \cite{goodfellow2014generative} are among the most widely used frameworks in image generation and have been extensively studied over the past few years. Conditional GANs (cGANs) \cite{mirza2014conditional} are variants of GANs that synthesize realistic and diverse images using conditional information, such as class labels. Depending on how the framework incorporates the data and class labels, most cGANs can be categorized into classifier-based cGANs \cite{odena2017conditional, gong2019twin, kang2020contragan, kang2021rebooting} and projection-based cGANs \cite{miyato2018cgans, miyato2018spectral, brock2018large, han2021dual}.
Classifier-based cGANs utilize conditional information (class labels) by training an additional classifier as well as a standard GAN discriminator. Meanwhile, projection-based cGANs propose a projection discriminator that takes an inner product between the embedded class labels and the feature vector extracted from the data.

The proposed method draws inspiration from all existing cGANs.
To the best of our knowledge, this is the first attempt to apply continuous conditional information to deblurring task.

\section{Preliminaries}
Generative Adversarial Networks (GANs) \cite{goodfellow2014generative} are well-established method for mimicking the probability distribution of the real data by playing a min-max game between the generator $G$ and discriminator $D$. Whereas $G$ learns to fool $D$ by generating realistic samples, $D$ learns to classify whether the given samples are true data (real) or generated data (fake). Their objective, $ V(G,D) $ is formulated as follows.
\begin{equation} \label{eq:vanilla_gan}
\begin{aligned}
    \mathop {\min }\limits_G \mathop {\max }\limits_D V(G,D)
    & = \mathbb{E}_{x \sim p(x)} [\log (D(x))] \\
    & + \mathbb{E}_{z \sim p(z)} [ \log (1-D(G(z)))],
\end{aligned}
\end{equation}
where $ p(x) $ denotes the real data distribution, and $p(z)$ denotes a pre-defined distribution, \eg, Gaussian distribution. A key property of GANs is that a well-trained $G$ successfully captures the data manifold even if there are missing data in the training set \cite{goodfellow2016nips, dosovitskiy2016learning, kumar2017semi}.

Conditional GANs (cGANs) \cite{mirza2014conditional, odena2017conditional, miyato2018cgans} are an extended GAN framework developed for conditional image synthesis. Given a pair of images $x$ and class labels $c$ sampled from the joint distribution of the real dataset $ (x,c) \sim p(x,c) $, the goal of $G$ is to learn the class-conditional image synthesis by utilizing $c$ as an additional input with $z$. Let $p_{G}(x|c)$ denote the generative distribution specified by $G(x,c)$ and $p_{G}(x,c):=p_{G}(x|c)p(c)$. The objective of generic cGANs \cite{mirza2014conditional}, $ V_{\text{cGAN}}(G,D) $, minimizes the Jensen-Shannon Divergence (JSD) between $p(x,c)$ and $p_{G}(x,c)$ as
\begin{equation} \label{eq:cgan}
\begin{aligned}
    \mathop {\min }\limits_G \mathop {\max }\limits_D 
    & V_{\text{cGAN}}(G,D) = \mathbb{E}_{(x,c) \sim p(x,c)} [\log (D(x,c))] \\
    & + \mathbb{E}_{z \sim p(z), c \sim p(c)} [\log(1-D(G(z,c), c))].
\end{aligned}
\end{equation}

As one of the most representative classifier-based cGANs, AC-GAN \cite{odena2017conditional} introduces an auxiliary classifier $Q$ to provide feedback on the class-conditional image synthesis of $G$. In AC-GAN, $D$ and $Q$ share all weights of the feature extractor, except for the final output layer. Let $p_Q(c|x)$ denote the conditional distribution induced by classifier $Q$. Then, their loss, $ V_{\text{AC-GAN}}(G,Q,D) $ can be expressed as follows
\begin{equation} \label{eq:AC_GAN}
\begin{aligned}
    & \mathop {\min }\limits_{G,Q} \mathop {\max }\limits_D V_{\text{AC-GAN}}(G,Q,D) = 
    \mathbb{E}_{(x,c) \sim p(x,c)} [\log (D(x))] \\
    & + \mathbb{E}_{z \sim p(z), c \sim p(c)} [\log(1-D(G(z,c)))] \\
    & -  \lambda_c \underbrace{\mathbb{E}_{(x,c) \sim p(x,c)} [\log (p_Q(c|x)]}_{\text{(a)}} \\
    & - \lambda_c \underbrace{\mathbb{E}_{(x,c) \sim p_G(x,c)} [\log (p_Q(c|x))]}_{\text{(b)}},
\end{aligned}
\end{equation}
where $\lambda_c$ is the balancing weight between the GAN and the auxiliary classification losses. In \cref{eq:AC_GAN}, the first two lines are loss functions similar to the original GANs (\cref{eq:vanilla_gan}), where $D$ serves as a binary classifier that distinguishes between real and fake samples. Terms $\text{(a)}$ and $\text{(b)}$ represent the auxiliary classification losses that enable $Q$ to determine the class labels of the input samples. Through this auxiliary classifier, AC-GAN can generate class-conditional image synthesis.

\begin{figure*}[t!]
    \centering
    \includegraphics[width=\linewidth]{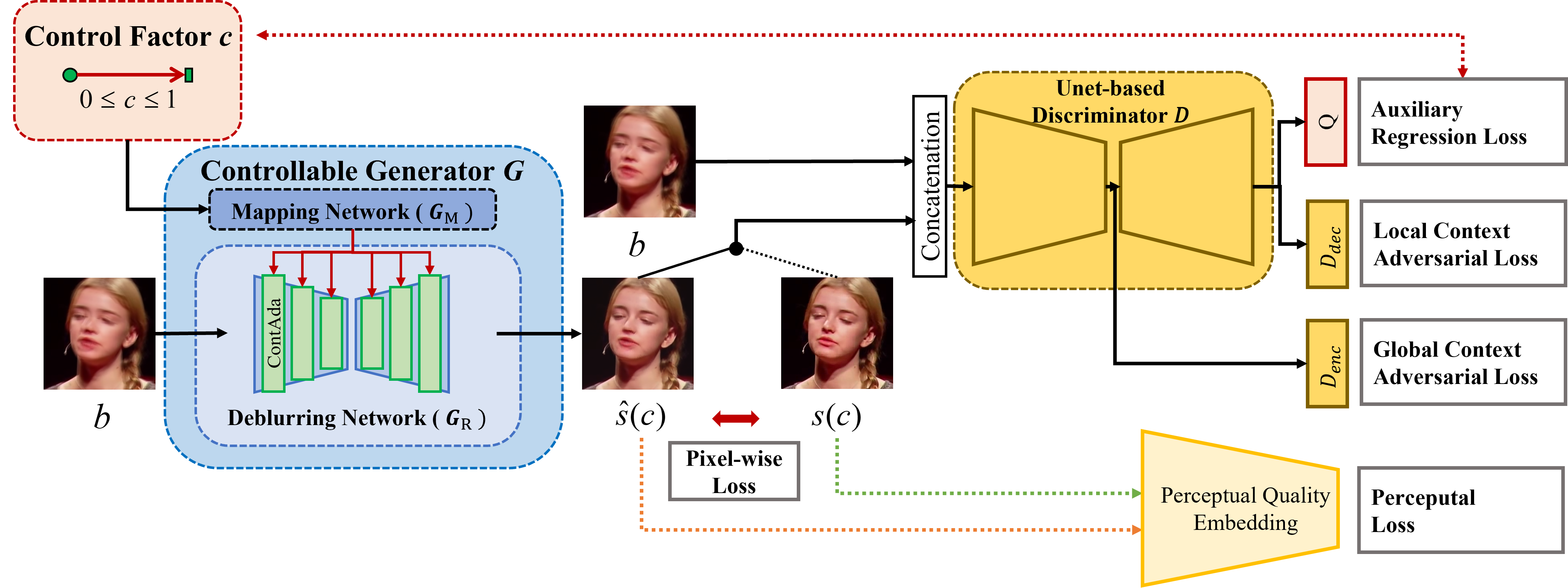}
    \caption{
    An overview of our CFMD-GAN framework.
    Given a single motion-blurred face image, the proposed generator restores the multiple sharp moments by varying a moment control factor.
    Subsequently, the proposed auxiliary regressor in the discriminator helps the generator learn to estimate more accurate result during training.
    }
    \label{fig:overview_cfmd}
\end{figure*}

\begin{figure}[t!]
    \centering
    \includegraphics[width=\linewidth]{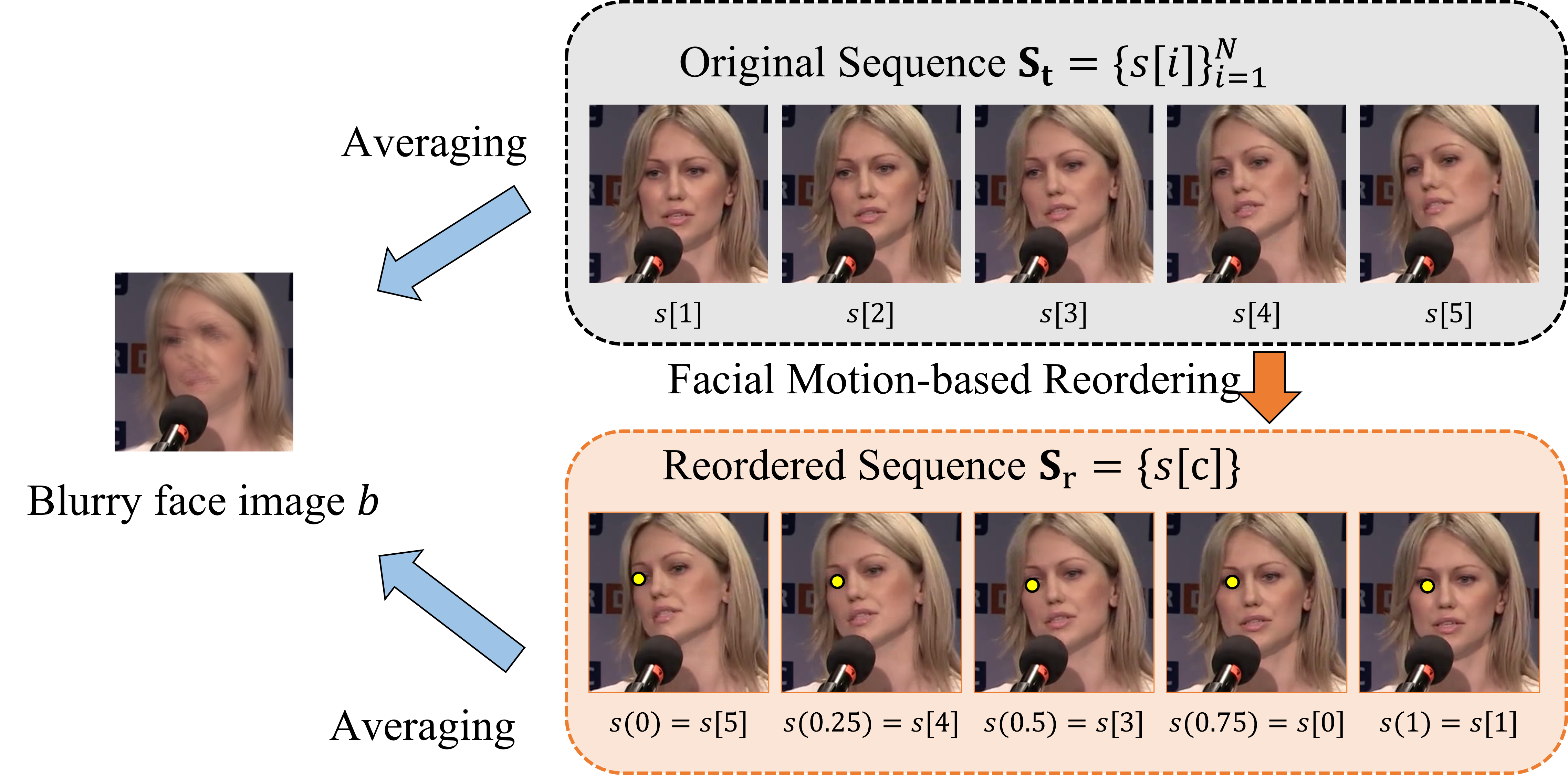}
    \caption{Facial motion-based reordering process (FMR).
    We rearrange the original sequence based on the position of the left eye \ie from top-left to right-bottom.
    }
    \label{fig:fmr}
\end{figure}

\begin{figure*}[t!]
    \centering
    \includegraphics[width=1.0\linewidth]{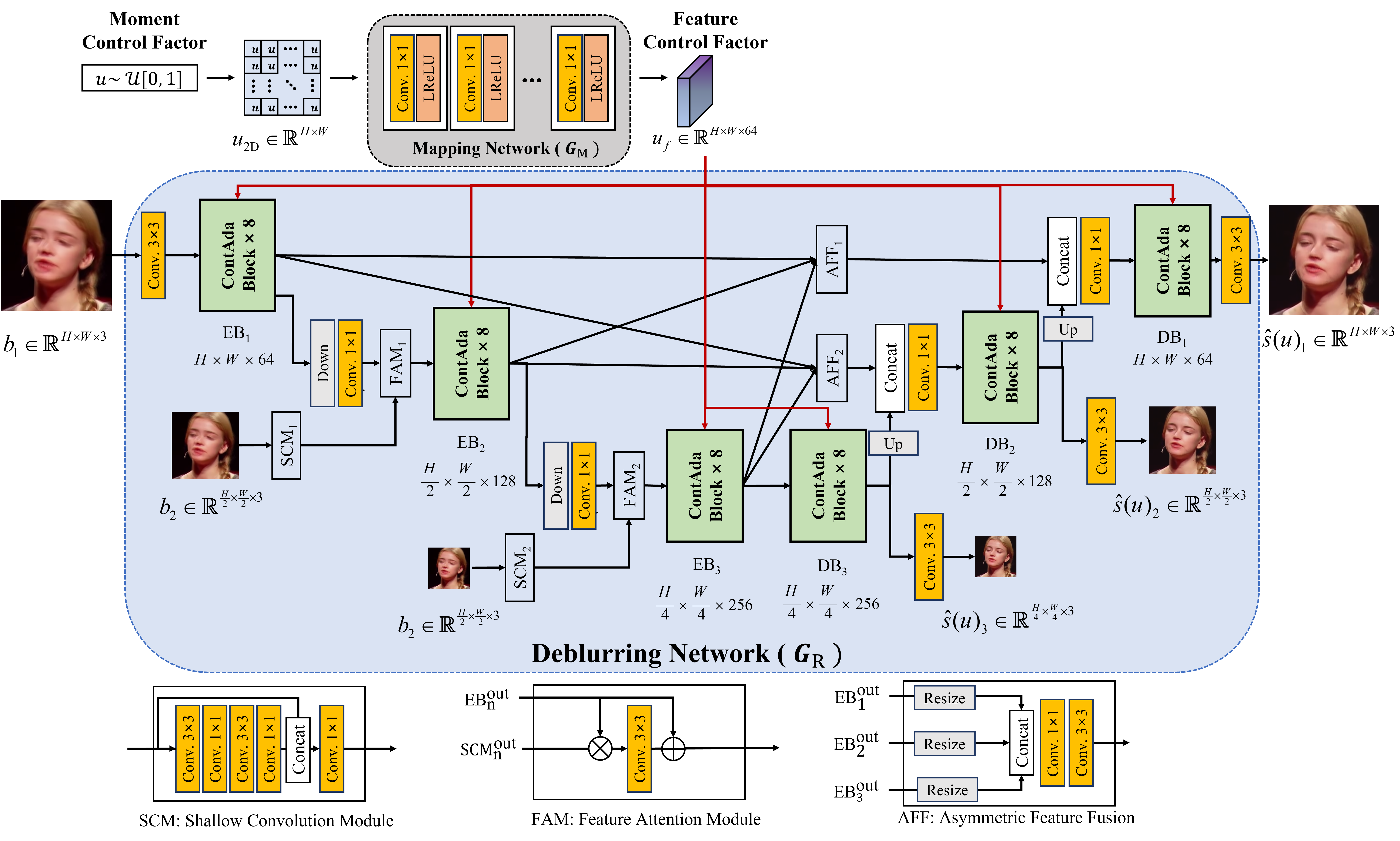}
    \caption{
    The architecture of our generator consisting of a mapping network and a deblurring network. 
    In the deblurring network, the proposed control-adaptive block incorporates features of control factors and features of blurred image.
    }
    \label{fig:cfmd_generator}
\end{figure*}

\section{Proposed Method} \label{sec:proposed_method}
In this section, we first introduce the facial motion-based reordering (FMR) process, which is proposed to mitigate the temporal ambiguity problem by utilizing human face information (\cref{sec:FMR}). Next, detailed explanation of the key components of the proposed CFMD-GAN is provided, which recovers the continuous moment latent in a blurry face image via a moment control factor (\cref{sec:CFMD_GAN}). Lastly, we introduce the training objectives of the proposed model (\cref{sec:model_objective}).

\subsection{Facial Motion-based Reordering} \label{sec:FMR}
One of the main challenges in restoring multiple images from a single blurred image is to resolve the \textit{temporal (sequence) ambiguity} of sharp moments. 
A motion-blurred image is the averaged result of a continuous sharp sequence during the exposure time \cite{hyun2013dynamic, Nah_2017_CVPR}. As averaging destroys the information of the temporal order \cite{jin2018learning, zhang2020every, argaw2021restoration}, reconstructing the original sequence of sharp moments is non trivial. 
For example, suppose a blurry facial image and its corresponding original sharp sequence are given, as shown in \cref{fig:fmr}. 
The problem is that the same blurry image can be obtained even if the face moves in a reverse or shuffled order during the exposure time. Owing to this ill-posed nature of the temporal ambiguity, finding the underlying sequence of the blurry image is one of the unsolved issues \cite{argaw2021restoration}. In this regard, previous studies \cite{jin2018learning, purohit2019bringing, argaw2021restoration} have found that temporal ambiguity causes unstable training of the network because it is difficult to uniquely define the temporal sequence of object movements.

To alleviate this, we leverage the information of the human face to apply effective yet strong constraints. In a recent study on face landmark detection, Sun \etal \cite{sun2019fab} proposed defining the intensity of facial motion as the movement of the left eye during the time unit. Inspired by this, we devised a facial motion-based reordering (FMR) that enables the network to restore sharp face images in a generalized order based on the position of the left eye.

Specifically, as depicted in \cref{fig:fmr}, FMR is a motion-based reordering process of the ground-truth (GT) sequence in a training dataset consisting of a single facial motion per single video clip. Let $ \mathbf{S_{t}} $ be a time-ordered set of GT frames sampled from a high-frame-rate facial video, which is denoted by 
\begin{equation}
    \mathbf{S_{t}} = \{ s[i] \in \mathbb{R}^{H \times W \times 3} \; | \; i \in [1, N] \},
\end{equation}
where $ i $ denotes the frame index within the total number of frames $N$. Then, a blurry image $ b \in \mathbb{R}^{H \times W \times 3} $ can be approximated by averaging these GT frames as follows: 
\begin{equation}
    b \simeq g( \frac{1} {N}\sum\nolimits_{i = 1}^{N} { s[i] } ),
\end{equation}
where $g( \cdot)$ denotes the camera response function \cite{Nah_2017_CVPR}.
We rearrange $ s[i] $ according to the position of the left eye $ ( x, y ) $ \footnote{ We utilize the public landmark detector provided by OpenCV \cite{opencv_library} to obtain the position of the left eye in the face image.} in each $ s[i] $ so that the frame that includes the eye in the top-left position comes first, and the frame that includes the eye in the bottom-right position follows the last.
Concretely, the proposed FMR process rearranges the sharp sequence according to the following criteria: 
\textbf{(c1)} The order is primarily determined by the ascending order of $ x $ values.
It generalizes the erratic movement of the face as a \textit{left-to-right movement}.
\textbf{(c2)} If there are frames with the same $ x $ values, those frames are sorted in ascending of $ y $ values.
It also regularizes the direction of facial motion to a \textit{top-to-bottom movement}.
\textbf{(c3)}  When frames have the same $ ( x, y ) $, they are sorted in ascending order of temporal sequence.

Following the above procedure, we further transform the frame index $i$ into a continuous motion index value $ u \in [0, 1] $ by applying $ u=\frac{i-1}{N} $. Then, we can denote this reordered set $ \mathbf{S_{r}} $ as follows:
\begin{equation}
    \mathbf{S_{r}} = \{ s(u) \in \mathbb{R}^{H \times W \times 3}  \; | \; u \in [0, 1] \}.
\end{equation}
Note that the real number $ u $ becomes a \textit{moment control factor} in the proposed framework.

In this study, the network learns to restore the facial motion-based order in $ \mathbf{S_{r}} $. It should be noted that this reordered sequence does not match the temporal sequence. Instead, the proposed framework restores all possible sharp moments latent in a blurry facial image. The FMR process allows the frames in the sequence $ \mathbf{S_{r}} $ to have regularity of face motion, which helps the network stabilize the training. The effects of the FMR are analyzed in \cref{sec:experiments}.

\subsection{Continuous Facial Motion Deblurring GAN} \label{sec:CFMD_GAN}
Inspired by the success of AC-GAN \cite{odena2017conditional}, the proposed continuous facial motion deblurring framework CFMD-GAN consists of a generator $G$ and a discriminator $D$ with an auxiliary regressor $Q$. An overview of the CFMD-GAN is depicted in \cref{fig:overview_cfmd}. Given a blurry face image and a control factor, $G$ performs the role of a deblurring network to perform conditional image restoration. Unlike most single image deblurring methods that only recover a single deblurred image from a single blurry image, the proposed $G$ is a function that restores a deblurred image conditioned on a control factor. That is, $G$ predicts continuous sharp moments latent in a blurry image by changing the value of the control factor. To achieve this, $D$ learns to predict 1) decisions of images belonging to real or fake \cite{schonfeld2020u} and 2) regression for control factor at the additional output layer $Q$.

\subsubsection{ Overall Pipeline of Generator}
Given a blurry face image $ b \in \mathbb{R}^{H \times W \times 3} $ and moment control factor $ u \in [0, 1] $ as the \textit{condition}, $G$ generates a restored face image $ \hat{s}(u) \in \mathbb{R}^{H \times W \times 3} $, which is defined as 
\begin{equation}
    \hat{s}(u) = G(b,u).
\end{equation}
Specifically, the proposed $G$ comprises two parts, a mapping network $G_{\text{M}}$ and a deblurring network $G_{\text{R}}$. First, $G_{\text{M}}$ translates the moment control factor $ u \in [0, 1] $ into the \textit{feature control factor} $ u_f \in \mathbb{R}^{H \times W \times 64} $. Second, $G_{\text{R}}$
incorporates $u_f$ with features extracted from $b$ and then outputs the final deblurred face image $ \hat{s}(u) $. 
In the proposed deblurring network, we deign a ContAda block so that $G$ can focus on important spatial locations and channels of features extracted from $b$ according to $c_f$.

\begin{figure*}[t!]
    \centering
    \includegraphics[width=1.0\linewidth]{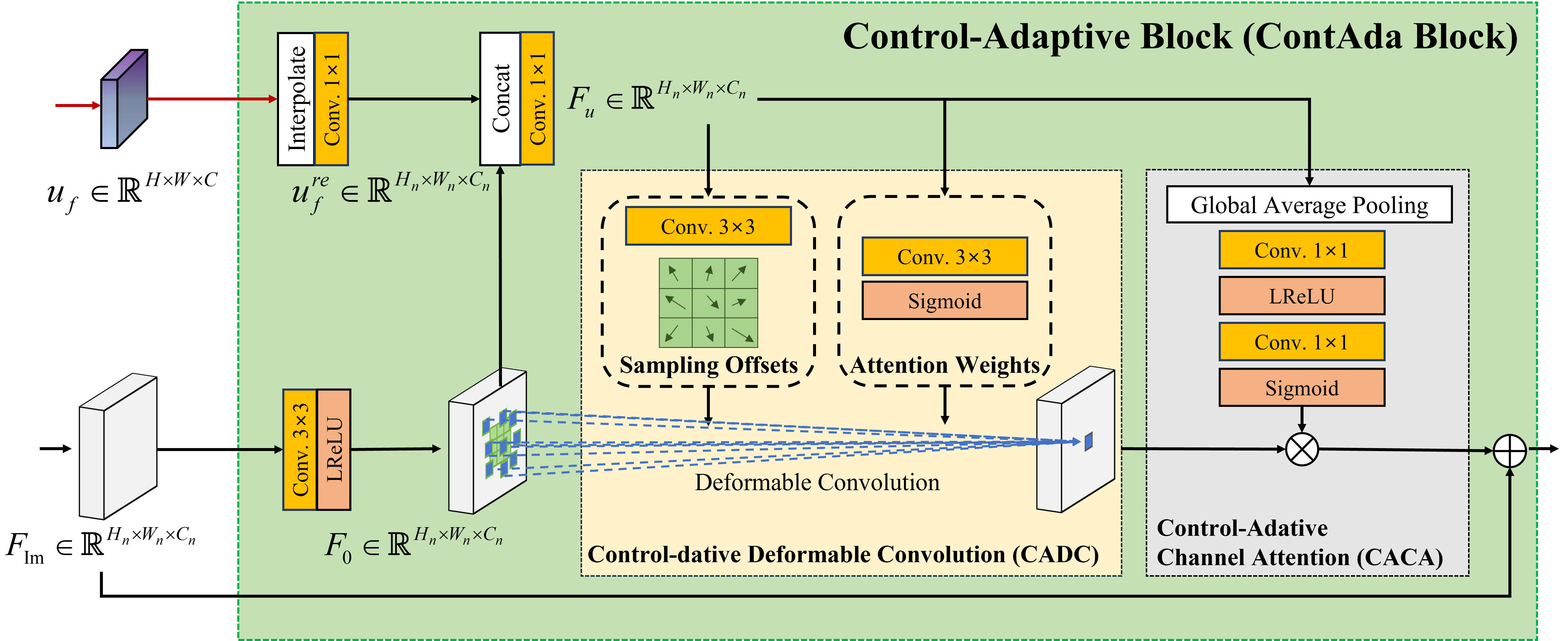}
    \caption{
        A structure of the proposed Control-Adaptive Block.
    }
    \label{fig:contada_block}
\end{figure*}

\noindent\textbf{ Mapping Network. }
In recent GANs studies \cite{karras2019style, shen2020interpreting, harkonen2020ganspace, zhu2020domain}, the additional mapping network has proven to provide more disentangled semantics for the generator than directly using input codes. Inspired by this, we set the mapping network  $G_{\text{M}} $ that outputs the \textit{ feature map control factor } $ u_f \in \mathbb{R}^{H \times W \times 64} $ from the given moment control factor $u \in [0, 1]$ as
\begin{equation}
    u_f = G_{\text{M}}(u).
\end{equation} 
As shown in \cref{fig:cfmd_generator}, $G_{\text{M}}$ first expands $ u $ into a 2-dimensional matrix $ u_{\text{2D}} \in \mathbb{R}^{H \times W} $ where each position is filled with $ u $. Then, $G_{\text{M}}$ outputs $ u_f $ from $ u_{\text{2D}} $ through several convolutional layers. Similar to \cite{karras2019style}, we design $G_{\text{M}}$ consisting of eight layers, each of which includes $ 1 \times 1 $ convolutions and a leaky ReLU \cite{maas2013rectifier}.

\noindent\textbf{ Deblurring Network. }
As mentioned earlier, the deblurring network $G_{\text{R}}$ generates a restored image $ \hat{s}(u) \in \mathbb{R}^{H \times W \times 3} $ from the blurry face image $b \in \mathbb{R}^{H \times W \times 3} $ and the feature map control factor $ u_f \in \mathbb{R}^{H \times W \times 64} $, as
\begin{equation}
    \hat{s}(u) = G_{\text{R}}(b, u_f).
\end{equation}
In this work, we employ the high-level structure of MIMO-UNet \cite{cho2021rethinking}, which has exhibited impressive performance in a single image deblurring field.
Specifically, as shown in \cref{fig:cfmd_generator}, MIMO-Unet is based on the encoder-decoder architecture and comprises three encoder blocks ($ \mathrm{EB_1}, \mathrm{EB_2}$ and $\mathrm{EB_3}$) and three decoder blocks ($ \mathrm{DB_1}, \mathrm{DB_2}$ and $\mathrm{DB_3}$). Each of these encoder and decoder blocks contain eight modified residual blocks \cite{tao2018scale}. 
Unlike the original MIMO-UNet, the network developed in this study can focus on important spatial positions and channels of the feature map depending on the control factor by replacing the residual blocks with the proposed ContAda blocks.
Note that SCM, FAM and AFF are modules used in the original MIMO-UNet that represent the shallow convolutional module, feature attention module and asymmetric feature fusion module, respectively. The details of each module, including the high-level architecture, can be found in \cite{cho2021rethinking}. In the following section, we discuss the proposed control-adaptive (ContAda) block.

\subsubsection{ Control-Adaptive Block }
There is a major challenge in applying existing building blocks (\eg variants of residual blocks \cite{he2016deep} ) that are widely used in single image deblurring networks in the proposed continuous facial motion deblurring. First, standard convolution-based layers have an inherent drawback in modelling geometric transformations. This drawback stems from the fact that a convolutional unit samples the input feature map at fixed spatial locations \cite{dai2017deformable, zhu2019deformable, zhu2020deformable}. To alleviate this, deformable convolution \cite{dai2017deformable, zhu2019deformable} has 
exhibited promising results in object detection by learning the offsets of the convolution grid to adjust the receptive field dynamically.
Inspired by this, several motion deblurring studies \cite{wang2019edvr, purohit2020region, yuan2020efficient} applied a deformable convolution module to handle the complex and various latent movements in a given blurred image \cite{wang2019edvr, purohit2020region}. 
However, these methods are still inadequate for our task because of the inability to focus on the adaptive positions of the feature maps depending on the control factor.

To this end, as shown in \cref{fig:contada_block}, we propose a Control-Adpative (ContAda) block that comprises a control-adaptive deformable convolution (CADC) module and a control-adaptive channel-attention (CACA) module. Let $F_{\mathrm{Im}} \in \mathbb{R}^{H_n \times W_n \times {C}_n } $ denote an input feature map of the ContAda block extracted from the input blurred image $b \in \mathbb{R}^{H \times W \times {3} } $. Here, $ H_n, W_n$ and  ${C}_n$ represent the height, width, and number of channels in the $n^{th}$ encoder/decoder block, respectively. The ContAda block starts with a $3 \times 3$ convolutional layer and LeakyReLU to extract the initial feature map $F_{o} \in \mathbb{R}^{H_n \times W_n \times {C}_n } $. Meanwhile, the feature control factor $u_f \in \mathbb{R}^{H \times W \times {C} } $, which is the output of the mapping network $ G_{\text{M}} $, is reshaped to $u^{(n)}_f \in \mathbb{R}^{H_n \times W_n \times {C}_n } $ using bilinear interpolation and $ 1 \times 1 $ convolutional layer.
Then, $u^{(n)}_f$ is concatenated with $ F_{o} $ along the channel dimension and then reshaped into $F_{u} \in \mathbb{R}^{H_n \times W_n \times {C}_n } $ by applying $ 1 \times 1 $ convolution layer. $F_{u}$ is utilized as an input feature for the CADC and CACA modules. In the following section, we introduce CADC and CACA distinctly.

\textit{Control-Adaptive Deformable Convolution} (CADC) module is based on deformable convolution \cite{dai2017deformable, zhu2019deformable} that enhances the ability of network in modeling spatial variations. Unlike \cite{dai2017deformable, zhu2019deformable}, where deformable offsets and attention weights are solely determined by internal information regarding the features of the input image, the proposed CADC learns the offsets and attention weights from the combined features of the control factor and image features. Let $K$ denote the sampling locations of a convolutional kernel. We denote the weight and pre-specified offset for the $k^{th}$ location as $w_k$ and $p_k$, respectively. For example, $ 3 \times 3 $ convolutional kernel of dilation 1 has $ 9 $ sampling locations ($K = 9$) and $ p_k \in \{ (-1,-1), (-1,0), \ldots ,(1,1) \} $. Let $ F_u (p) $ and $ F_{dc} (p) $ denote the features at location $p$ of the input feature map $F_u$ and output feature map $ F_{dc} $, respectively. Accordingly, the proposed CADC can be formulated as
\begin{equation}
\begin{aligned}
    F_{dc}(p) = \sum\limits_{k = 1}^K {w_k  \cdot F_{u} ( p + p_k  + \Delta p_k ) } \cdot \Delta m_k,
\end{aligned}
\end{equation}
where $ \Delta p_k $ and $ \Delta m_k $ denote the learned offset and attention weight scalar for the $k^{th}$ location, respectively. As shown in \cref{fig:contada_block}, $ \Delta p_k $ and $ \Delta m_k $ are determined by separate convolutional layers. The output of the sampling offsets branch has $2K$ channels, corresponding to $ \{ \Delta p_k \}^{K}_{k=1} $. The output of the attention weights branch is of $K$ channels, as $ \{ \Delta m_k \}^{K}_{k=1} $, and each $ \Delta m_k $ is in the range of $[0,1]$ by the sigmoid function. Following \cite{zhu2019deformable}, the initial values of $ \Delta p_k $ and $ \Delta m_k $ are set to 0 and 0.5, respectively.

\textit{Control-Adaptive Channel Attention} (CACA) module is mainly motivated by \cite{chen2017sca, hu2018squeeze, zhang2018image}, which benefits from applying the channel-wise attention mechanism for convolutional layers. In short, both CADC and CACA can be considered as attention functions of two variables: features extracted from blurry images and those extracted from the control factor. They are complementary in that CADC performs spatial attention to select important geometric properties of features, whereas CACA focuses on significant semantic and contextual attributes \cite{chen2017sca, zhang2018image}. Given $F_u$, as can be seen in \cref{fig:contada_block}, global average pooling is applied to transform channel-wise information into channel descriptors, following \cite{zhang2018image}. Subsequently, we obtain the channel-wise attention weights from two $ 1 \times 1 $ convolutional layers and a sigmoid function. The learned attention weights are multiplied by $F_{dc}$, the output of the CADC, in an element-wise manner.

\subsubsection{ Discriminator } \label{subsec:discriminator}
As shown in \cref{fig:overview_cfmd}, the proposed discriminator $D$ is based on the U-net structure discriminator \cite{schonfeld2020u} with an auxiliary regressor. 
In our framework, $G$ receives as inputs a blurred face image $b$ and a control factor $u$, and outputs an image $\hat{s}(u) = G(b,u) $. Following \cite{isola2017image}, the discriminator $D$ takes as inputs as a blurred face image and the corresponding sharp face image. Here, a face image is either a real sharp image ${s}(u)$ drawn from the training dataset or a restored image $\hat{s}(u)$ from $G$. Then, $D$ provides three types of outputs from the encoder output layer $D_{enc}$, decoder output layer $D_{dec}$, and auxiliary regression layer $Q$.

Following \cite{schonfeld2020u}, $D_{enc}$ determines whether the global input context is real or fake. Similarly, the final outputs of $D_{dec}$ are used to classify whether the local context of the input is sampled from the real or fake. On the other hand, the proposed $Q$ provides a regression value for the estimated control factor. Instead of predicting a single scalar value of $c$, our $Q$ outputs $\hat{u}_{2D} \in \mathbb{R}^{H \times W}$ and is trained to estimate the ground-truth control factor ${u}_{2D} \in \mathbb{R}^{H \times W}$.

\subsection{Model Objectives} \label{sec:model_objective}
Following \cite{goodfellow2014generative}, $D$ and $G$ are optimized alternately using loss functions, which are described as follows.

\subsubsection{Discriminator Loss}
To estimate the global and per-pixel probability distributions, the encoder loss $\mathcal{L}_{D_{enc}}$ and decoder loss $\mathcal{L}_{D_{dec}}$ are formulated as follows:
\begin{equation} \label{eq:dis_enc_loss}
\begin{aligned}
\mathcal{L}_{D_{enc}} = 
    & - \log D_{enc}(b, s(u))  + \log D_{enc}(b, G(b,u)),
    \\
    \mathcal{L}_{D_{dec}} = 
    & \frac{1}{WH} \sum\limits_{i,j}^{W,H} \Big( -\log [D_{dec}(b, s(u))]_{(i,j)} \\
    & + \log [ D_{dec}(b, G(b,u)) ]_{(i,j)} \Big).
\end{aligned}
\end{equation}
Here, $[D_{dec}(\cdot)]_{(i,j)}$ represents the decision of the discriminator decoder at pixel coordinate $(i,j)$.

To ensure that the restored image is an accurate moment of the blurry image, the auxiliary regression loss $\mathcal{L}_{Q}$ is defined by
\begin{equation} \label{eq:gen_info_loss}
\begin{aligned}
\mathcal{L}_{Q} =  \frac{1}{WH} \sum\limits_{i,j}^{W,H} \Big(
 & \left\| { u_{\text{2D}}  - Q(b, s(u)) } \right\|_2^2 \\
 + & \left\| { u_{\text{2D}}  - Q(b, G(b,u)) } \right\|_2^2 \Big)
\end{aligned}
\end{equation}

The total loss of $D$ is formulated as the sum of the above objectives:
\begin{equation} \label{eq:dis_total_loss}
\mathcal{L}_{D} = \mathcal{L}_{D_{enc}} + \mathcal{L}_{D_{dec}} + \lambda_Q \mathcal{L}_Q,
\end{equation}
where $\lambda_Q$ denotes a weight parameter, which is empirically set to 0.05.

\subsubsection{Generator Loss}
\noindent\textbf{Auxiliary Regression Loss.}
To accurately restore the output image conditioned by the control factor, an auxiliary regression loss $\mathcal{L}_{ar}$ is optimized as follows:
\begin{equation} \label{eq:gen_ar_loss}
\begin{aligned}
    \mathcal{L}_{ar} = \frac{1}{WH} \sum\limits_{i,j}^{W,H} \left\| { u_{\text{2D}}  - Q(b, G(b,u)) } \right\|_2^2.
\end{aligned}
\end{equation}

\noindent\textbf{Adversarial Loss.}
We use the Unet-discriminator to ensure that the generated image is indistinguishable from the real data for both global and local contexts. 
The adversarial loss $\mathcal{L}_{adv}$ is formulated as follows:
\begin{equation} \label{eq:gen_adv_loss}
    \begin{aligned}
    \mathcal{L}_{adv} = - \Big(
    & \log D_{enc}(b, G(b, u)) \\
    & + \frac{1}{WH} \sum\limits_{i,j}^{W,H} \log [ D_{dec}(b, G(b, u))]_{(i,j)} \Big).
    \end{aligned}
\end{equation}

\noindent\textbf{Pixel-wise Loss.}
To restore accurate pixel intensities, following \cite{zamir2021multi}, we employ the Charbonnier loss \cite{charbonnier1994two} to minimize the pixel-wise distance between a ground-truth moment and a restored image as follows:
\begin{equation} \label{eq:gen_pix_loss}
\mathcal{L}_{pix} = \sum\limits_{n = 1}^3 {\frac{1}
{{W_n H_n}}} \sum\limits_{i,j}^{W_n,H_n}
    \sqrt {\left\| { s(u)_n - G(b,u)_n \ } \right\|^2  + \varepsilon ^2 },
\end{equation}
where $n$ denotes the number of multi-scale levels. $H_n$ and $W_n$ represent the height and width at the corresponding $n^{th}$ level of output image, respectively. 
Following \cite{zamir2021multi}, $ \varepsilon $ is set to $ {10}^{-3} $.

\noindent\textbf{Perceptual loss.}
Furthermore, we use perceptual loss to obtain perceptually satisfactory images. Similar to \cite{jo2020investigating}, LPIPS \cite{zhang2018unreasonable} is employed for perceptual loss.
\begin{equation} \label{eq:gen_per_loss}
\mathcal{L}_{per} = 
\sum\limits_{l}^{M} {\omega ^l 
    \left\| {\phi ^l ( s(u) ) -  \phi ^l ( G(b,u) )}  \right\|_2^2}
\end{equation}
Here, $\phi (\cdot)$ is a feature extractor, $ \omega $ denotes a learned vector to measure the LPIPS score, and the total score is averaged over $M$ layers.

In overall, the total loss of $G$ combines the aforementioned loss functions,
\begin{equation} \label{eq:gen_total_loss}
    \mathcal{L}_{G} = 
        \lambda_{ar} \mathcal{L}_{ar} + \lambda_{adv} \mathcal{L}_{adv} + \lambda_{pix} \mathcal{L}_{pix} + \lambda_{per} \mathcal{L}_{per},
\end{equation}
where $\lambda_{ar}, \lambda_{adv}, \lambda_{pix}$ and $\lambda_{per}$ denote the balancing weights empirically set to 0.05, 0.1, 1 and 0.01, respectively.

\section{Experiments} \label{sec:experiments}
\subsection{Experimental Setup}
\subsubsection{Dataset}
We use the 300VW dataset \cite{shen2015first} which consists of a large number of high-quality facial videos recorded in the wild. Each video has a duration of about one minute at 25-30 fps. 
Following the face deblurring study by Ren \etal \cite{ren2019face}, the training and test datasets are extracted from 83 videos and 9 videos, respectively.
Each blurry image is synthesized by averaging various numbers (5-13) of consecutive sharp frames, as in recent motion deblurring studies \cite{Nah_2017_CVPR, ren2019face}.
Thus, the testset consists of total 13,058 blurred images and 116,188 sharp frames.
The details of the number of test images are listed in \Cref{tab:testdataset}.

\begin{table}
    \caption{
            Configuration of facial motion deblurring testset synthesized using 300VW dataset \cite{shen2015first}.
            }
    \label{tab:testdataset}
    \renewcommand{\arraystretch}{1.2}
    \begin{adjustbox}{width=1\linewidth}
    \centering
    \begin{tabular}{c|c|c}
    \noalign{\smallskip}\noalign{\smallskip}
    \toprule
        $\#$ of averaged frames & $\#$ of blurred images & $\#$ of sharp images \\
        \hline
        5 & 2753 & 13765 \\
        7 & 2677 & 18739 \\
        9 & 2605 & 23445 \\
        11 & 2530 & 27830 \\
        13 & 2493 & 32409 \\
        \hline
        Total & 13058 & 116188 \\
    \bottomrule
    \end{tabular}
    \end{adjustbox}
\end{table}

\subsubsection{Implementation Details}
The proposed framework is implemented with Pytorch \cite{paszke2019pytorch} and trained with NVIDIA TITAN-RTX GPUs. 
We train our networks using the Adam optimizer \cite{kingma2014adam} with ${\beta}_1 $ = 0.9, and ${\beta}_2 $ = 0.999. 
The initial learning rate is set as $1 \times 10^{-4}$ and it decayed exponentially by a factor of 0.99 for every epoch. 
For data augmentation, we randomly scale the image from 1.0 to 1.5 and then randomly crop the image with a spatial size of $256 \times 256 \times 3$. During training, we set the batch size as 8 and train our model for 200 epochs.

\subsubsection{Evaluation Metrics}
For a quantitative evaluation, we measure the PSNR and SSIM \cite{wang2004image}, which are traditionally used for image quality assessment.
We also report two metrics of learning-based perceptual quality, FID \cite{heusel2017gans} and LPIPS \cite{zhang2018unreasonable}.
Moreover, we employ the ArcFace \cite{deng2019arcface} model to measure the distance of facial identity between the ground truth (GT) and the resulting image, as \cite{wang2021towards}.

\begin{table}
    \caption{ 
        Quantitative comparison of single-to-single general deblurring methods. 
        The best and the second best results are highlighted in \textbf{bold} and \underline{underline}, respectively. 
        }
    \label{tab:quan_s2s_general}
    \renewcommand{\arraystretch}{1.2}
    \begin{adjustbox}{width=1\linewidth}
    \centering
    \begin{tabular}{l|ccccc}
    \noalign{\smallskip}\noalign{\smallskip}
    \toprule
        Methods & PSNR ($ \uparrow $ ) & SSIM ($ \uparrow $ ) & LPIPS ($ \downarrow $ ) & FID ($ \downarrow $ ) & ArcFace ($ \downarrow $ ) \\
        \hline
        Nah \etal \cite{Nah_2017_CVPR} & 31.4144 & 0.9232 & 0.0935 & 13.9722 & 1.1250 \\
        SRN \cite{tao2018scale} & 32.1485 & 0.9249 & 0.0930 & 11.6292 & 1.1488 \\
        DMPHN \cite{Zhang_2019_CVPR} & 33.1797 & 0.9284 & 0.0847 & 13.0407 & 1.0338  \\
        DMPHN* & {33.8182} & {0.9345} & 0.0916 & 14.3071 & 1.0126 \\
        MIMO \cite{cho2021rethinking} & 34.0372 & 0.9350 & 0.0795 & 7.8606 & 1.0205 \\
        MIMO* & \textbf{34.8496} & \textbf{0.9401}  & \underline{0.0794} & \underline{7.2459}  & \underline{0.9918} \\
        \hline
        CFMD-GAN & \underline{34.2684} & \underline{0.9362} & \textbf{0.0697} & \textbf{5.1448} & \textbf{0.9338} \\
    \bottomrule
    \end{tabular}
    \end{adjustbox}
\end{table}

\begin{figure*}[t!]
    \captionsetup[subfloat]{labelformat=empty}
    \centering
    \subfloat[]{
        \includegraphics[width=0.18\linewidth]{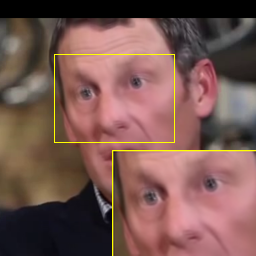}
    }\hfill \hspace*{-0.5em}  \vspace*{-1.5em} 
    \subfloat[]{
        \includegraphics[width=0.18\linewidth]{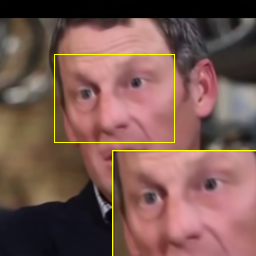}
    }\hfill \hspace*{-0.5em}   
    \subfloat[]{
        \includegraphics[width=0.18\linewidth]{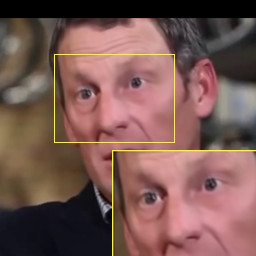}
    }\hfill \hspace*{-0.5em}   
    \subfloat[]{
        \includegraphics[width=0.18\linewidth]{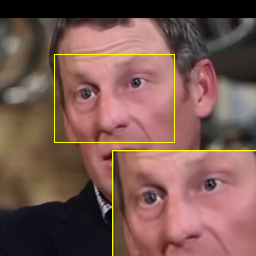}
    }\hfill \hspace*{-0.5em}  
    \subfloat[]{
        \includegraphics[width=0.18\linewidth]{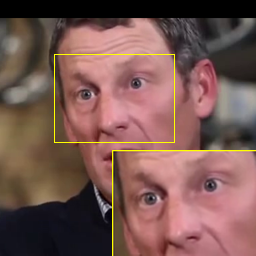}
    }\hfill \hspace*{-0.5em}  
    \\
    \subfloat[Input]{
        \includegraphics[width=0.18\linewidth]{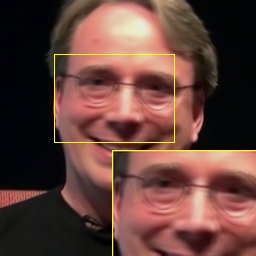}
    }\hfill \hspace*{-0.5em}   
    \subfloat[DMPHN* \cite{Zhang_2019_CVPR}]{
        \includegraphics[width=0.18\linewidth]{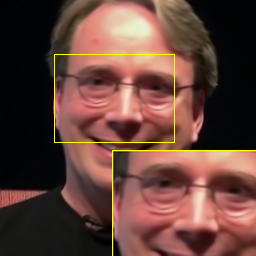}
    }\hfill \hspace*{-0.5em}   
    \subfloat[MIMO* \cite{cho2021rethinking}]{
        \includegraphics[width=0.18\linewidth]{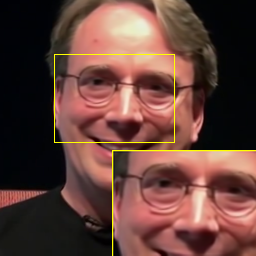}
    }\hfill \hspace*{-0.5em}   
    \subfloat[CFMD-GAN (ours)]{
        \includegraphics[width=0.18\linewidth]{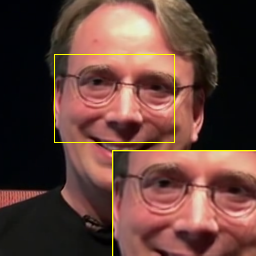}
    }\hfill \hspace*{-0.5em}  
    \subfloat[Ground truth]{
        \includegraphics[width=0.18\linewidth]{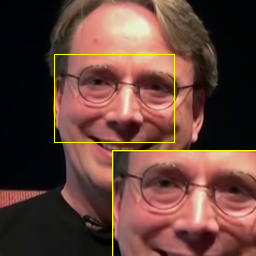}
    }\hfill \hspace*{-0.5em}  
 \caption{
    Qualitative comparisons of single-to-single general deblurring methods.
    Zoom in for the best view.
    }
\label{fig:qual_s2s_general}
\end{figure*}

\subsection{Comparisons with the state-of-the-arts}
To the best of our knowledge, the proposed method is the first attempt for single-to-video face deblurring.
Hence, we conduct extensive and faithful comparisons with state-of-the-art methods in single image deblurring.
Specifically, the proposed CFMD-GAN is compared with single-to-single (s2s) general deblurring ( \ie Nah \etal \cite{Nah_2017_CVPR}, SRN \cite{tao2018scale}, DMPHN \cite{Zhang_2019_CVPR}, MIMO \cite{cho2021rethinking}), s2s face deblurring (\ie Shen \etal \cite{shen2018deep}, UMSN \cite{yasarla2020deblurring}, MSPL \cite{lee2020progressive} ), and single-to-video (s2v) general deblurring ( \ie Jin \etal \cite{jin2018learning}).
To facilitate fair comparisons, we retrain the existing methods using the same training dataset used in this study.
The retrained models are marked with asterisks (*).
All experiments are performed using the official codes provided by the authors.

\subsubsection{Single-to-Single General Deblurring}
In this comparison, we evaluate the performance of the center frame prediction, as most s2s general methods are proposed to restore the center frame. 
For the proposed method, the control factor is set to $ c = 0.5 $ to obtain the center frame results.
\Cref{tab:quan_s2s_general} reports the comparisons of s2s general deblurring methods. 
Despite the significant improvements in the performance of retrained DMPHN* and MIMO* compared to the original DMPHN and MIMO, our CFMD-GAN shows the best results in LPIPS, FID and ArcFace distance, and the second best in PSNR and SSIM.

As investigated in recent GAN-based restoration studies \cite{ledig2017photo, blau20182018, wang2018esrgan, chen2018fsrnet, jo2020investigating, wang2021towards, gu2021ntire}, PSNR and SSIM may be lower because the GAN-based model tends to generate fake yet realistic details and textures \cite{gu2021ntire}.
This effect of GANs can be clearly observed in the visual comparisons in \cref{fig:qual_s2s_general}.
Compared with other methods, the proposed CFMD-GAN restores more realistic textures and finer details of facial components, such as the eyes, nose, and eyelids.
Based on these results, we can confirm that the proposed model can predict a more accurate center frame than the other methods. 

\begin{table}
    \caption{
        Quantitative comparison of single-to-single face deblurring methods.
        The best and the second best results are highlighted in \textbf{bold} and \underline{underline}, respectively.
        }
    \label{tab:quan_s2s_face}
    \renewcommand{\arraystretch}{1.2}
    \begin{adjustbox}{width=1\linewidth}
    \centering
    \begin{tabular}{l|ccccc}
    \noalign{\smallskip}\noalign{\smallskip}
    \toprule
        Methods & PSNR ($ \uparrow $ ) & SSIM ($ \uparrow $ ) & LPIPS ($ \downarrow $ ) & FID ($ \downarrow $ ) & ArcFace ($ \downarrow $ ) \\
        \hline
        Shen \etal \cite{shen2018deep} & 23.1795 & 0.6873  & 0.2310 & 78.3630 & 2.2112 \\
        UMSN \cite{yasarla2020deblurring} & 27.0050 & 0.8276 & 0.1460 & 39.4150 & 1.4908 \\
        UMSN* & {30.5884}  & {0.9140}  & {0.0833}  & {16.8517} & {1.2716} \\
        MSPL-GAN \cite{lee2020progressive} & 28.2286 & 0.8936 & 0.1092 & 27.9441 & 1.2947 \\
        MSPL-GAN* & \textbf{34.2711} & \underline{0.9359} & \underline{0.0638} & \underline{10.3597} & \underline{1.0983} \\
        \hline
        CFMD-GAN$_{128}$ & \underline{33.8475}  & \textbf{0.9379} & \textbf{0.04910} & \textbf{6.5449}  & \textbf{1.0246}  \\
    \bottomrule
    \end{tabular}
    \end{adjustbox}
\end{table}

\begin{figure*}[t!]
    \captionsetup[subfloat]{labelformat=empty}
    \centering
    \subfloat[]{
        \includegraphics[width=0.16\linewidth]{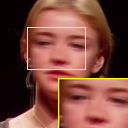}
    }\hfill \hspace*{-0.5em}  \vspace*{-1.5em} 
    \subfloat[]{
        \includegraphics[width=0.16\linewidth]{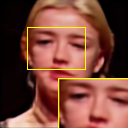}
    }\hfill \hspace*{-0.5em} 
    \subfloat[]{
        \includegraphics[width=0.16\linewidth]{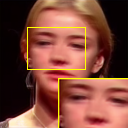}
    }\hfill \hspace*{-0.5em}  
    \subfloat[]{
        \includegraphics[width=0.16\linewidth]{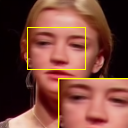}
    }\hfill \hspace*{-0.5em}  
    \subfloat[]{
        \includegraphics[width=0.16\linewidth]{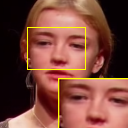}
    }\hfill \hspace*{-0.5em}  
    \subfloat[]{
        \includegraphics[width=0.16\linewidth]{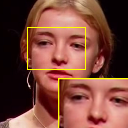}
    }\hfill \hspace*{-0.5em}  
    \\
    \subfloat[Input]{
        \includegraphics[width=0.16\linewidth]{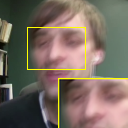}
    }\hfill \hspace*{-0.5em}   
    \subfloat[Shen \etal \cite{shen2018deep}]{
        \includegraphics[width=0.16\linewidth]{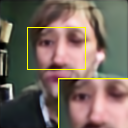}
    }\hfill \hspace*{-0.5em}   
    \subfloat[UMSN* \cite{yasarla2020deblurring}]{
        \includegraphics[width=0.16\linewidth]{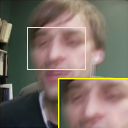}
    }\hfill \hspace*{-0.5em}  
    \subfloat[MSPL-GAN* \cite{lee2020progressive}]{
        \includegraphics[width=0.16\linewidth]{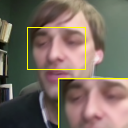}
    }\hfill \hspace*{-0.5em}  
    \subfloat[ CFMD-GAN$_{128}$ (ours) ]{
        \includegraphics[width=0.16\linewidth]{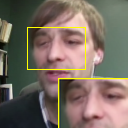}
    }\hfill \hspace*{-0.5em}   
    \subfloat[Ground truth]{
        \includegraphics[width=0.16\linewidth]{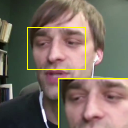}
    }\hfill \hspace*{-0.5em}  
\caption{
    Qualitative comparisons of single-to-single face deblurring methods.
    Zoom in for the best view.
    }
\label{fig:qual_s2s_face}
\end{figure*}

\subsubsection{Single-to-Single Face Deblurring}
Most existing s2s face deblurring methods \cite{shen2018deep, yasarla2020deblurring, lee2020progressive} are developed to remove spatially-uniform blurs.
However, our training and test datasets contain spatially-variant blurs.
Besides, their models only handle input images of $128 \times 128 \times 3$. 
For these reasons, we downsample our dataset to $128 \times 128 \times 3$ and use it to retrain UMSN \cite{yasarla2020deblurring}, MSPL \cite{lee2020progressive} and our model (termed as CFMD-GAN$_{128}$).
The retrained models, UMSN* and MSPL*, are trained to predict the center frame, similar to the s2s general deblurring approaches.
Note that we do not retrain Shen \etal \cite{shen2018deep} because they do not release the training code.

\Cref{tab:quan_s2s_face} and \cref{fig:qual_s2s_face} provide the quantitative and qualitative comparisons of the s2s face deblurring methods, respectively.
In this experiment, the proposed method achieved significantly better performance on SSIM, LPIPS, FID and ArcFace than the existing face deblurring methods. For PSNR, our method achieved the second best.
Shen \etal \cite{shen2018deep} fails to restore plausible results because they are not trained to remove spatially-variant blurs, as shown in \cref{fig:qual_s2s_face}.
Although the retrained models (UMSN* and MSPL*) show improved performance, they are still inferior to the CFMD-GAN.

\begin{table}
    \caption{
            Quantitative comparison of single-to-video deblurring methods.
            \enquote{\# of GT} indicates the number of GT frames per a single blurry image, \enquote {\# of pairs} is the total number of test GT frames, and \enquote{ALL} represents the entire results of 300VW testset. 
            Note that all the results of CFMD-GAN are measured with the same model.
            The best results are highlighted in \textbf{bold}.
            }
        \label{tab:quan_s2v_general}
    \renewcommand{\arraystretch}{1.2}
    \begin{adjustbox}{width=1\linewidth}
    \centering
    \begin{tabular}{l|ccccccc}
    \noalign{\smallskip}\noalign{\smallskip}
    \toprule
        Methods & $\#$ of GT & $\#$ of pairs  & PSNR ($ \uparrow $ ) & SSIM ($ \uparrow $ ) & LPIPS ($ \downarrow $ ) & FID ($ \downarrow $ ) & ArcFace ($ \downarrow $ ) \\
        \hline
        Jin \etal \cite{jin2018learning} & 7 & 18739 & 29.2407 & 0.8754 & 0.1471 & 25.6946 & 1.1574 \\
        CFMD-GAN & 7 & 18739 & \textbf{33.1360} & \textbf{0.9336}  & \textbf{0.0691} & \textbf{3.4238}  & \textbf{0.9078} \\
        \hline
        CFMD-GAN & 5 & 13765 & 34.6556 & 0.9498 & 0.0538  & 2.9080 & 0.7548 \\
        CFMD-GAN & 9 & 23445 & 32.0300 & 0.9192 & 0.0823 & 4.0663 & 1.0384 \\
        CFMD-GAN& 11 & 27830 & 31.1256 & 0.9060 & 0.0939 & 4.7120 & 1.1466 \\
        CFMD-GAN& 13 & 32409 & 30.3970 & 0.8949 & 0.1041 & 5.3640 & 1.2367 \\
        \hline
        CFMD-GAN & ALL & 116188 & 31.8474 & 0.9153 & 0.0857 & 4.0948 & 1.0650 \\
    \bottomrule
    \end{tabular}
    \end{adjustbox}
\end{table}

\begin{figure*}[t!]
    \captionsetup[subfloat]{labelformat=empty}
    \subfloat[]{
        \includegraphics[width=0.18\linewidth]{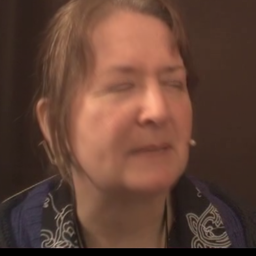}
    }\hfill \hspace*{-0.5em} \vspace*{-1.5em} 
    \subfloat[]{
        \animategraphics[loop, autoplay, width=0.18\linewidth]{5}{images/comparisons_video/158_blur007_002879/gt/frame-}{0}{6}
    }\hfill \hspace*{-0.5em} 
    \subfloat[]{
        \animategraphics[loop, autoplay, width=0.18\linewidth]{5}{images/comparisons_video/158_blur007_002879/jin/frame-}{0}{6}
    }\hfill \hspace*{-0.5em}  
    \subfloat[]{
        \animategraphics[loop, autoplay, width=0.18\linewidth]{5}{images/comparisons_video/158_blur007_002879/cfmd_07/frame-}{0}{6}
    }\hfill \hspace*{-0.5em}   
    \subfloat[]{
        \animategraphics[loop, autoplay, width=0.18\linewidth]{5}{images/comparisons_video/158_blur007_002879/cfmd_51/frame-}{0}{50}
    }\hfill \hspace*{-0.5em}
    \\
    \subfloat[Input]{
        \includegraphics[width=0.18\linewidth]{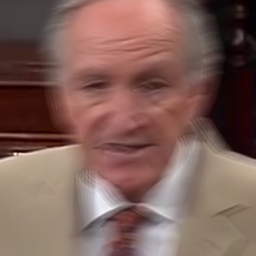}
    }\hfill \hspace*{-0.5em}  
    \subfloat[ GT ]{
        \animategraphics[loop, autoplay, width=0.18\linewidth]{5}{images/comparisons_video/522_blur007_000019/gt/frame-}{0}{6}
    }\hfill \hspace*{-0.5em} 
    \subfloat[ Jin \etal \cite{jin2018learning} ]{
        \animategraphics[loop, autoplay, width=0.18\linewidth]{5}{images/comparisons_video/522_blur007_000019/jin/frame-}{0}{6}
    }\hfill \hspace*{-0.5em}  
    \subfloat[ CFMD-GAN 7 frames ]{
        \animategraphics[loop, autoplay, width=0.18\linewidth]{5}{images/comparisons_video/522_blur007_000019/cfmd_07/frame-}{0}{6}
    }\hfill \hspace*{-0.5em}   
    \subfloat[ CFMD-GAN 51 frames ]{
        \animategraphics[loop, autoplay, width=0.18\linewidth]{5}{images/comparisons_video/522_blur007_000019/cfmd_51/frame-}{0}{50}
    }\hfill \hspace*{-0.5em}
    \caption{
    Qualitative comparisons of single-to-video deblurring methods.
    \textit{This figure contains videos that are best viewed using Adobe Reader.}
    }
    \label{fig:qual_s2v_general}
\end{figure*}

\subsubsection{Single-to-Video General Deblurring}
For s2v general deblurring methods, we compare our method with Jin \etal \cite{jin2018learning} which officially released their test model.
Since this method is strictly fixed to extract seven sequential frames from a single blurry image, we compare the results only for blurry images averaged by seven sharp frames.
None of the s2v deblurring methods \cite{jin2018learning, purohit2019bringing, zhang2020every, argaw2021restoration} have released their training codes. \cite{jin2018learning} is the only work that provides the test code.

\Cref{tab:quan_s2v_general} reports quantitative comparisons with Jin \etal \cite{jin2018learning} and detailed results of our model according to the number of GT frames.
The model of Jin \etal \cite{jin2018learning} is limited to predicting only a fixed number of frames when the model is trained once.
However, it is worth to note that the proposed single model can predict various numbers of output frames without additional network changes or training processes. Visual comparisons in \cref{fig:qual_s2v_general} show this difference.

\begin{figure}[t!]
    \captionsetup[subfloat]{labelformat=empty}
   \subfloat[]{
        \includegraphics[width=0.4\linewidth]{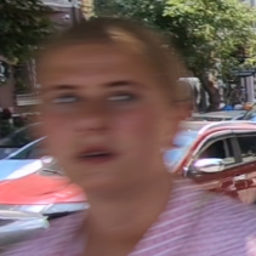}
    }\hfill \hspace*{-0.5em}  \vspace*{-1.5em} 
    \subfloat[]{
        \animategraphics[loop, autoplay, width=0.4\linewidth]{5}{images/REDS_result/00000098/cfmd_11/frame-}{0}{10}
    }\hfill \hspace*{-0.5em} 
    \\
    \subfloat[]{
        \includegraphics[width=0.4\linewidth]{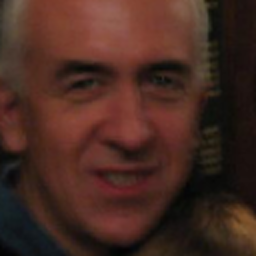}
    }\hfill \hspace*{-0.5em} \vspace*{-1.5em}
    \subfloat[]{
        \animategraphics[loop, autoplay, width=0.4\linewidth]{5}{images/Lai_result/16/cfmd_11/frame-}{0}{10}
    }\hfill \hspace*{-0.5em} 
    \\
    \subfloat[ Input ]{
        \includegraphics[width=0.4\linewidth]{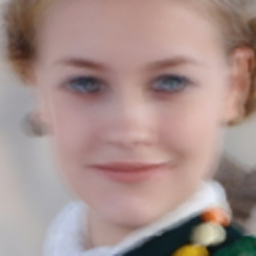}
    }\hfill \hspace*{-0.5em} 
    \subfloat[ CFMD-GAN 11 frames ]{
        \animategraphics[loop, autoplay, width=0.4\linewidth]{5}{images/Lai_result/11/cfmd_11/frame-}{0}{10}
    }\hfill \hspace*{-0.5em}  
    \\
    \caption{Qualitative results of the proposed CFMD-GAN on REDS dataset \cite{Nah_2019_CVPR_Workshops_REDS} ($1^{st}$ row) and Lai dataset \cite{lai2016comparative} ($2^{nd}$ and $3^{rd}$ rows). 
    \textit{This figure contains videos that are best viewed using Adobe Reader.}}
    \label{fig:other_dataset}
\end{figure}

\subsection{Analysis on CFMD-GAN}
\subsubsection{Evaluation on Other Test Datasets}
Since our model is trained and evaluated with synthetically blurred images using the 300VW dataset \cite{shen2015first}, we verify how our model performs on other motion-blur benchmark datasets such as REDS \cite{Nah_2019_CVPR_Workshops_REDS} and Lai \etal \cite{lai2016comparative}.
The REDS dataset is generated using 120 fps videos, synthesizing blurry frames by merging consecutive frames.
The Lai dataset contains real-blur images where the GT images do not exist.
We manually crop the facial regions of images in the REDS validation set and the Lai dataset. 

\cref{fig:other_dataset} shows that our method restores satisfactory images for recent benchmark deblurring datasets.
In $1^{st}$ row of \cref{fig:other_dataset}, we can see that our method produces not only a sharp face, but also the background that was occluded by the face in the previous frame.  
For the real-blurred images in $2^{nd}$ row of \cref{fig:other_dataset}, our model restores plausible results containing consecutive frames. 
Our framework can provide all sharp moments that user wants from a single motion-blurred face image.

\begin{table}
    \caption{
            Ablations on the proposed ContAda block.
            The best results are highlighted in \textbf{bold}.
            }
    \label{tab:quan_ours_ablation_ContAda}
    \renewcommand{\arraystretch}{1.2}
    \begin{adjustbox}{width=1\linewidth}
    \centering
    \begin{tabular}{c|c|c|ccccc}
    \noalign{\smallskip}\noalign{\smallskip}
    \toprule
         CADC & CACA & FMR & PSNR ($ \uparrow $ ) & SSIM ($ \uparrow $ ) & LPIPS ($ \downarrow $ ) & FID ($ \downarrow $ ) & ArcFace ($ \downarrow $ ) \\
        \hline \hline
        \checkmark &  & \checkmark & 33.5546 & 0.9263 & 0.0814 & 7.0030 & 1.0457 \\
        & \checkmark & \checkmark & 34.0271 & 0.9279 & 0.0810 & 6.8322 & 1.0389 \\
        \hline 
        \checkmark & \checkmark & & 33.4478 & 0.9201 & 0.0880 & 9.6110 & 1.1547 \\
        \hline
        \checkmark & \checkmark & \checkmark &\textbf{34.2684} & \textbf{0.9362} & \textbf{0.0697} & \textbf{5.1448} & \textbf{0.9338} \\ 
    \bottomrule
    \end{tabular}
    \end{adjustbox}
\end{table}

\subsubsection{Ablation Study}
In \Cref{tab:quan_ours_ablation_ContAda}, we evaluate the impact of the proposed ContAda block consisting of ContAda deformable convolution (CADC) and ContAda channel attention (CACA).
With the CADC module, the proposed method can focus on the spatially important sampling points of the feature maps by the feature map control factor.
Notably, using only CACA module improves the average PSNR by about 0.5dB
compared to using only CADC module.
This demonstrates that the channel attention plays a more important role in the proposed model. 
More importantly, using both CADC and CACA achieves the best results.
This indicates that both spatial and channel-wise modulations are required for the continuous facial motion deblurring.
Furthermore, we conduct an ablation study to investigate the contribution of FMR to the network training.
The $3^{rd}$ row in \Cref{tab:quan_ours_ablation_ContAda} indicates that without FMR, the performance of the model drops drastically when it learns the original temporal order. 

\section{Conclusion}
In this study, we introduce CFMD-GAN, a novel framework for continuous facial motion deblurring with a single network and a single training process.
Subsequently, we apply facial motion-based reordering (FMR) to mitigate the difficulty of temporal ordering by utilizing domain-specific facial information.
This ensures a stable learning process for the framework.
We devise an auxiliary regressor to learn continuous motion deblurring by integrating the concept of conditional GANs into a single image deblurring framework.
In addition, we propose a control-adaptive (ContAda) block that focuses on deformable locations and important channels according to the control factor.
In our extensive experiments, we demonstrate that the proposed method outperforms state-of-the-art methods in facial image deblurring. 
The proposed framework can provide continuous sharp moments that users want to obtain from a single motion-blurred facial image.
Since the proposed method restores facial motion in the order of FMR, there may be a limitation in predicting the accurate temporal order of the facial motion.
However, we believe that the proposed method will be the basis for future studies on continuous facial motion deblurring. In addition, incorporating various facial priors can be a fundamental issue for future research to improve the quality of this study.

{\small
\bibliographystyle{ieee_fullname}
\bibliography{references}
}

\end{document}